\documentclass[sigconf,natbib=true,anonymous=false]{acmart}

\setcopyright{none}
\copyrightyear{}
\acmYear{}
\acmDOI{}
\acmISBN{}
\acmConference{}{}{}
\acmBooktitle{}
\AtBeginDocument{%
  }
\usepackage[table]{xcolor}
\usepackage{pifont}

\begin{document}

\title{LAUKIN: A Multi-jurisdictional Common Law  Contract Dataset}


\author{Amrita Singh}
\affiliation{
  \institution{Computer Science and Engineering}
  \city{UNSW, Sydney}
  \country{Australia}
}
\email{amrita.singh1@unsw.edu.au}

\author{Aditya Joshi}
\affiliation{
  \institution{Computer Science and Engineering}
  \city{UNSW, Sydney}
  \country{Australia}
}
\email{aditya.joshi@unsw.edu.au}

\author{Jiaojiao Jiang}
\affiliation{
  \institution{Computer Science and Engineering}
  \city{UNSW, Sydney}
  \country{Australia}
}
\email{jiaojiao.jiang@unsw.edu.au}

\author{Hye-young Paik}
\affiliation{
  \institution{Computer Science and Engineering}
  \city{UNSW, Sydney}
  \country{Australia}
}
\email{h.paik@unsw.edu.au}

\author{May Fong Cheong}
\affiliation{
  \institution{Law and Justice}
  \city{UNSW, Sydney}
  \country{Australia}
}
\email{mf.cheong@unsw.edu.au}

\renewcommand{\shortauthors}{Singh et al.}

\begin{abstract}
Multinational companies increasingly require cross-jurisdictional contract review, yet existing legal NLP datasets are largely restricted to a single jurisdiction. We introduce \textbf{LAUKIN} (\textbf{L}egal equivalence dataset of \textbf{A}ustralia, \textbf{UK}, and \textbf{IN}dia), a dataset of clause pairs (AU-UK, UK-IN, IN-AU) labelled for boolean legal equivalence. We develop a novel multi-stage retrieval and reranking pipeline to construct the initial clause pair mapping,  with a subset of clause pairs subsequently annotated by legal experts as Equivalent or Not Equivalent. The dataset comprises 14,727 clause pairs from 204 contracts across 8 agreement types, of which 3,000 are manually labelled: 900 train, 600 dev, and 1,500 test. We evaluate 12 models across 4 techniques, achieving a best macro-F1 of 65.11\%, establishing LAUKIN as a challenging benchmark. Results reveal that, despite shared legal heritage, drafting conventions diverge significantly across jurisdictions, making cross-jurisdictional equivalence classification non-trivial. LAUKIN also includes 11,727 unlabelled training pairs to support future semi-supervised learning research in legal NLP.
\end{abstract}



\maketitle

\section{Introduction}
\label{0}
Natural language processing (NLP) for the legal domain has witnessed remarkable progress over the past decade, with applications spanning contract review, judgment prediction and summarization, statutory interpretation, and information retrieval from case law \cite{singh2025survey, ariai2025natural, akter2025comprehensive, feng2024legal}. These advancements are reshaping legal practice for scholars, educators, researchers, and practitioners by automating time-intensive processes, cutting operational costs, and reducing human error \cite{singh2025survey, ariai2025natural, kluttz2019automated, huang2024optimizing, ucheagwu2025legal}. However, existing datasets and benchmarks remain largely single-jurisdictional, restricted to specific countries such as the USA, China, and the European Union, limiting their applicability and coverage across other jurisdictions \cite{singh2025survey}. This is problematic in an increasingly globalised world, where multinational companies not only need to adapt their legal contracts to the distinct terminology, clause structures, and enforceability standards of different jurisdictions, but also perform cross-jurisdictional contract review. The challenge is amplified among jurisdictions such as Australia, the UK, and India, which are rooted in the same English common law tradition, but their contract laws have diverged significantly through independent court systems, distinct legislation, and local legal conventions \cite{barnett2024reflections, kumar2022contract}. Consequently, their clauses differ in lexical choices, syntactic and pragmatic structure, making cross-jurisdictional contract review a complex task. For instance, consider an equivalent force majeure clause pair:
\begin{quote}
\small
\textit{India:} Neither Party shall be responsible for delays or failures in performance resulting from acts of god, acts of civil or military authority, fire, flood, strikes, war, epidemics, shortage of power, or other acts or causes reasonably beyond the control of such Party.

\textit{Australia:} Neither party will be liable for any delay or non-performance of its obligations under this Agreement to the extent that the delay or non-performance is caused or contributed to by a Force Majeure Event, subject to compliance with this clause 16.
\end{quote}

This pair illustrates an instance of lexical variation: the Indian clause enumerates disaster events such as floods, epidemics, and shortage of power, whereas the Australian clause adopts the  term Force Majeure Event. The absence of multi-jurisdictional datasets for legal NLP impedes the automation of such tasks across common law jurisdictions.
\begin{figure*}[t]
  \centering
\includegraphics[width=0.86\linewidth]{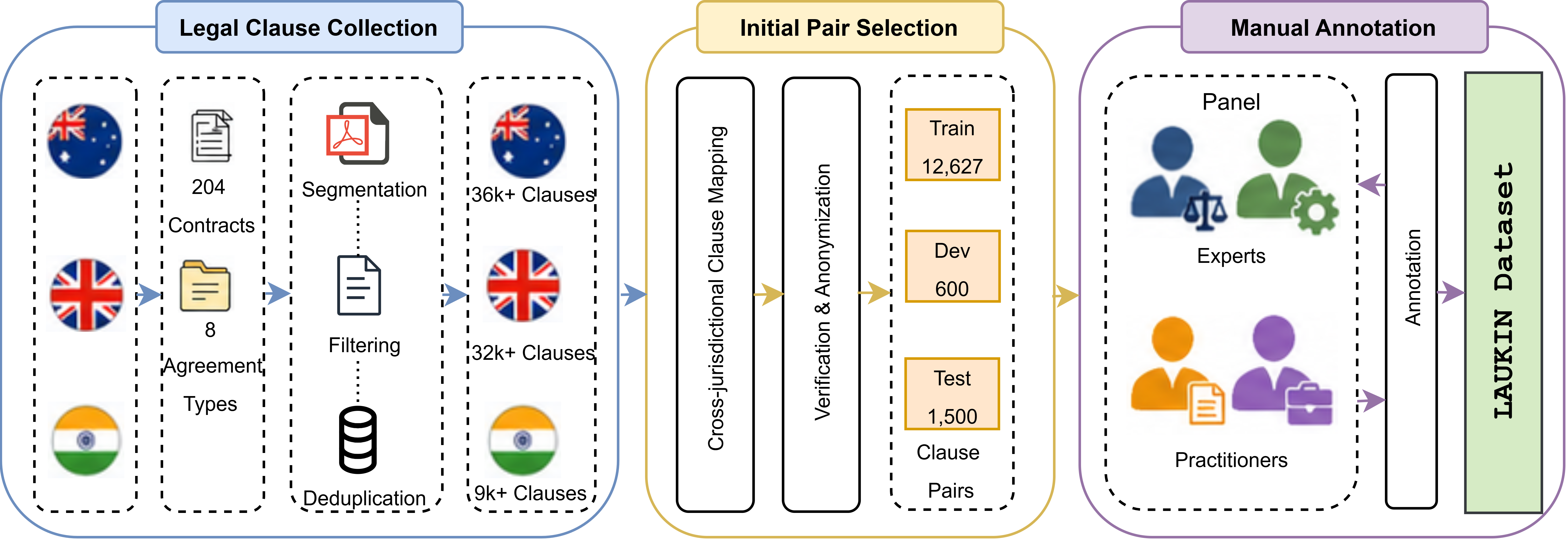}
  \caption{Creation of LAUKIN: Legal Clause Collection (\S\ref{2.1}), 
Initial Pair Selection (\S\ref{2.2}), and Manual Annotation (\S\ref{2.3})}
  \label{fig_4}
\end{figure*}
To address this gap, we introduce \textbf{LAUKIN} (\textbf{L}egal equivalence dataset of \textbf{A}ustralia, \textbf{UK}, and \textbf{IN}dia), the \textbf{first multi-jurisdictional common law contract dataset} consisting of paired English legal clauses: AU-UK, UK-IN, and IN-AU. LAUKIN is constructed semi-automatically via a novel multi-stage retrieval and reranking pipeline for cross-jurisdictional clause mapping, with a subset of clause pairs annotated by legal experts as Equivalent or Not Equivalent. The dataset comprises 14,727 clause pairs extracted from 204 contracts across 8 agreement types. Of these, 3,000 are manually labelled and split into 900 train, 600 dev, and 1,500 test sets. We evaluate 12 models across 4 techniques, with the best macro-F1 of 65.11\%, demonstrating that LAUKIN serves as a challenging benchmark. In addition to the labelled data, we also provide 11,727 unlabelled training clause pairs as a resource for future semi-supervised learning research in legal NLP. LAUKIN enables applications including cross-jurisdictional contract review and clause drafting, legal information retrieval, and LLM benchmarking. The LAUKIN dataset and all associated code (covering both, dataset creation and benchmarking) will be made publicly available upon acceptance.

\section{Creation of LAUKIN}
Figure \ref{fig_4} illustrates the three-stage creation workflow: Legal Clause Collection, Initial Pair Selection, and Manual Annotation.

\subsection{Legal Clause Collection}
\label{2.1}
Contracts for Australia are sourced from \href{https://www.tenders.gov.au/}{AusTender}\footnote{\url{https://www.tenders.gov.au/}, \href{https://creativecommons.org/licenses/by/3.0/au/}{CC BY 4.0 AU}, accessed 30 Nov 2025.}, for the UK from \href{https://www.contractsfinder.service.gov.uk/}{Contract Finder}\footnote{\url{https://www.contractsfinder.service.gov.uk/}, \href{http://www.nationalarchives.gov.uk/doc/open-government-licence/version/3}{OGL v3.0}, accessed 30 Nov 2025.}, and for India from various publicly accessible central and state government portals. All extracted contracts are freely available for public use; the majority are sample or model contracts serving as agreement templates, along with a few expired
contracts.
The collection process involves manual verification of source correctness and reliability, review of licensing policies across over 100 government state and territory websites, and removal of duplicate contracts, incurring 7 person-days of effort. Finally, a total of 204 contracts are selected across Australia (67), the UK (83), and India (54), covering 8 agreement types as detailed in Table \ref{tab_1}. These contracts vary widely in length, ranging from a few pages to 900 pages. UK and Australia contracts reach up to 900 and 750 pages respectively, whereas India contracts are comparatively shorter, with a maximum length of 250 pages.
\begin{table}[ht!]
\centering
\caption{Distribution of Contracts and Resultant Clauses}
\resizebox{0.42\textwidth}{!}{
\begin{tabular}{lccc}
\toprule
\textbf{Contract/Agreement Type} & \textbf{Australia} & \textbf{UK} & \textbf{India} \\
\midrule
Service & 7 & 8 & 5 \\
Rental \& Lease & 7 & 11 & 10 \\
Franchise & 6 & 7 & 2 \\
Loan & 4 & 2 & 4 \\
Marketing \& Consultancy & 8 & 9 & 1 \\
Partnership, NDA \& Employment & 7 & 20 & 16 \\
Construction \& Maintenance & 17 & 12 & 3 \\
Hosting \& License & 11 & 14 & 13 \\
\midrule
\textbf{Total Contracts (204)} & \textbf{67} & \textbf{83} & \textbf{54} \\
\textbf{Total Clauses (w/o Post-processing)} & \textbf{50,111} & \textbf{47,058} & \textbf{13,844} \\
\midrule
\textbf{Total Clauses (w Post-processing)} &\textbf{36,942} &\textbf{32,716} & \textbf{9,183}\\
\midrule
\end{tabular}
}
\label{tab_1}
\end{table}
Clauses are automatically extracted from each contract PDF using \texttt{PyMuPDF} 
(\texttt{fitz}). Raw text is parsed page by page and segmented into sentences via 
regex-based boundary detection. Sentences shorter than 10 words are discarded to 
eliminate headers, footers, and fragments. This yields a raw corpus of over 50K, 
47K, and 13K clauses for Australia, the UK, and India, respectively. Since contracts within the same jurisdiction frequently reuse identical or near-identical clauses, deduplication is applied independently within each jurisdiction using \texttt{rapidfuzz}, where a character 4-gram blocking index is constructed to avoid an $O(n^2)$  all-pairs comparison.
Pairwise similarity is computed via 
\texttt{fuzz.ratio}, and any two clauses scoring $\geq 90\%$ are grouped into a 
duplicate cluster. Within each cluster, only the longest clause is retained. This 
per-jurisdiction deduplication reduces the final corpus to 36K+, 32K+, and 9K+ 
clauses for Australia, the UK, and India, respectively.

\subsection{Initial Pair Selection}
\label{2.2}
\begin{figure}[t]
  \centering
\includegraphics[width=0.48\textwidth]{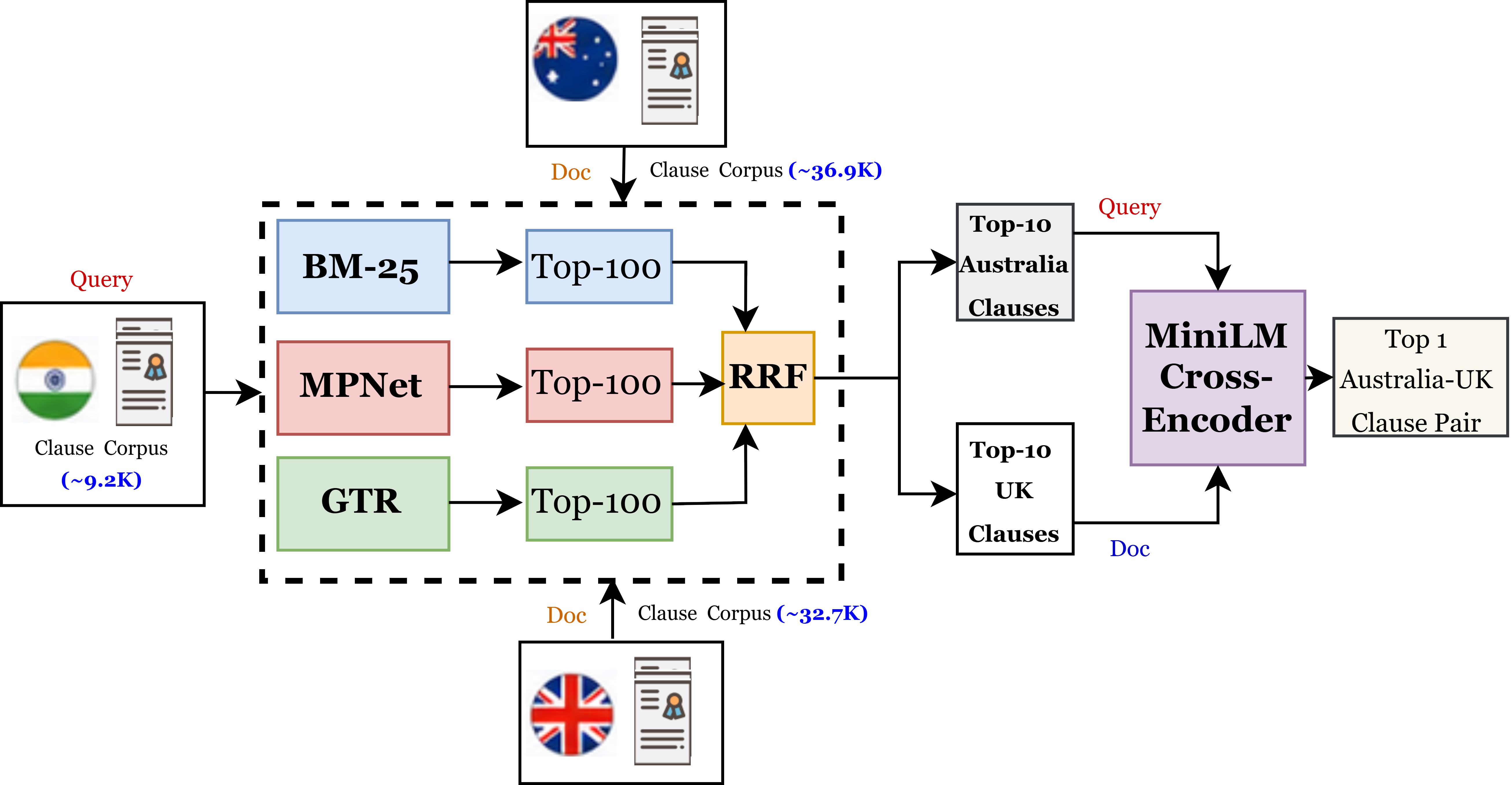}
  \caption{Cross-jurisdictional Clause Mapping Pipeline}
  \label{fig_2}
\end{figure}

We propose a novel multi-stage retrieval and reranking pipeline for cross-jurisdictional clause mapping. The clause corpus for each jurisdiction, collected in the previous step, serves as input to the pipeline, as shown in Figure \ref{fig_2}. As India has the fewest clauses (9,183) compared to Australia and the UK, each Indian clause serves as a query and is matched against the most lexically and semantically equivalent clause from both the Australian and UK corpora. For each corpus, the top-100 candidates are fetched by three retrievers in parallel: BM25 for lexical matching, and MPNet (\texttt{all-mpnet-base-v2}) and GTR (\texttt{gtr-t5-base}) for semantic retrieval. The resulting ranked lists are merged using Reciprocal Rank Fusion (RRF) \cite{cormack2009reciprocal} into a single fused candidate list per corpus. This results in the top-10 Australian and top-10 UK candidate clauses for each Indian query clause. These candidates are then reranked using a Cross-Encoder (\texttt{ms-marco-MiniLM-L-6-v2}), where Australian candidates serve as queries and UK candidates serve as documents. The Australian-UK pair with the highest reranking score is selected as the best match for the Indian query clause, yielding a clause triplet (IN-AU-UK). Once a triplet is formed, each matched clause is removed from its respective corpus, ensuring unique triplet mapping. The final output is a set of 9,183 IN-AU-UK clause triplets. For ethical and legal reasons, all personally identifiable information is anonymised, replacing organisation names, places, and email addresses with \texttt{<Organization>}, \texttt{<Place>}, and \texttt{<Email>} respectively. Next, a qualitative check is performed, as contracts vary in structure, formatting, and drafting style across agreement types and jurisdictions, and may contain incomplete or non-clause segments that convey general information rather than contractual obligations. Any triplet containing such segments is removed entirely to ensure data quality. The anonymisation and qualitative check are carried out manually by an annotator, requiring 5 person-days. Finally, 4,909 high-quality triplets remain and are split into three sets: train (4,209 triplets), dev (200 triplets), and test (500 triplets). Each triplet in these sets is decomposed into three clause pairs (AU-UK, UK-IN, IN-AU), resulting in 12,627 train, 600 dev, and 1,500 test clause pairs, totalling 14,727 clause pairs.
\subsection{Manual Annotation}
\label{2.3}
For the dataset annotation and task formulation, a panel of two experts and two practitioners is formed. The law expert has 30+ years of experience specialising in contract and commercial law. The NLP expert has 14+ years of experience in dataset creation and methodology development for NLP. Both experts guide the labelling instructions and task design. The two practitioners are a Legal NLP researcher with 4+ years of experience in contract annotation and requirement implementation, and an Associate Lawyer at a law firm specialising in corporate and commercial law. As labelling the entire dataset requires both legal expertise and significant time, a semi-supervised annotation setting is adopted, reflecting the real-world challenge where fully annotated datasets are rarely available. A clause pair is labelled as Equivalent if both clauses share the same core legal function and clause type, irrespective of lexical or syntactic variation. A clause pair is labelled as Not Equivalent if the clauses differ in legal function or clause type, resulting in different legal obligations or outcomes.  The two practitioners independently annotate the 600 dev clause pairs. Inter-Annotator Agreement (IAA) is computed, yielding Cohen's $\kappa = 0.75$ (substantial agreement \cite{landis1977measurement}). Disagreements are resolved through iterative rounds of discussion between the experts and practitioners. Following completion of the dev set, the 1,500 test and 900 train clause pairs are annotated by the second practitioner, resulting in a high-quality labelled dataset. The total annotation effort for LAUKIN amounts to 25 person-days.

\section{LAUKIN Statistics}
LAUKIN consists of 14,727 clause pairs. The clause pairs are split into 12,627 train (900 labelled, 11,727 unlabelled), 600 labelled development, and 1,500 labelled test sets, as shown in Table \ref{tab_2}. The dataset is imbalanced, with Equivalent clause pairs forming the minority class, accounting for only 10.6\%, 9.7\%, and 26.7\% of the train, dev, and test splits respectively. This reflects a real-world challenge, where clauses across jurisdictions differ more than they align due to each country's distinct legal system and requirements \cite{kumar2022contract, saxena2023comparative}. We further analyse the word overlap between clause pairs using Jaccard similarity (J). Across all splits, the vast majority of pairs exhibit low word overlap (J < 0.25), and even among Equivalent pairs, only 18.9\%, 25.9\%, and 15.0\% of train, dev, and test pairs have J $\geq$ 0.25. This suggests that lexical overlap alone is insufficient to determine equivalence. 
\begin{table}[ht!]
\centering
\caption{LAUKIN Dataset Statistics. Lab. = Labelled; Unlab. = Unlabelled; \%HS = percentage of pairs with Jaccard $\geq 0.25$}
\label{tab_2}
\resizebox{0.98\columnwidth}{!}{%
\begin{tabular}{l crr rr rr}
\toprule
 & \multicolumn{3}{c}{\textbf{Train}} & \multicolumn{2}{c}{\textbf{Dev}} & \multicolumn{2}{c}{\textbf{Test}} \\
\cmidrule(lr){2-4} \cmidrule(lr){5-6} \cmidrule(lr){7-8}
\textbf{Class} & \textbf{Unlab.} & \textbf{Lab.} & \textbf{\%HS} & \textbf{Lab.} & \textbf{\%HS} & \textbf{Lab.} & \textbf{\%HS} \\
\midrule
Equivalent     & -- & 95  & 18.9\% & 58  & 25.9\% & 400  & 15.0\% \\
Not Equivalent & -- & 805 & 6.3\%  & 542 & 6.8\%  & 1100 & 4.5\%  \\
\midrule
\textbf{Total (14,727)} & \textbf{11,727} & \textbf{900} & \textbf{7.7\%} & \textbf{600} & \textbf{8.7\%} & \textbf{1,500} & \textbf{7.3\%} \\
\bottomrule
\end{tabular}
}
\end{table}
Next, we present examples from LAUKIN to illustrate both classes. The Equivalent clause pair in Section \ref{0} shows an Indian and an Australian force majeure clause that exempt parties from liability for delays beyond their control, despite lexical and syntactic differences. In contrast, the Not Equivalent clause pair shown below shares 36\% word overlap but differs in legal function: the Australian clause governs direct service delivery by the Consultant, while the UK clause governs personnel provision by the Consultant Company to the Council.
\begin{quote}
\small
\textit{Australia:} The Consultant has agreed to provide the Services and any Additional Services in accordance with the terms and conditions of this agreement.

\textit{UK:} The Consultant Company shall make available to the Council the Individual to provide the Services on the terms of this agreement.
\end{quote}
\section{Experimental Setup}
We introduce the legal equivalence classification task using LAUKIN. Given a clause pair, a model predicts whether the pair is Equivalent or Not Equivalent. We evaluate 12 models across four  baseline techniques: Zero-Shot \cite{kojima2022large}, Few-Shot \cite{brown2020language}, CoT prompting \cite{wei2022chain}, and SFT \cite{ouyang2022training}. For prompting-based techniques, 8 decoder-based models are used via OpenRouter\cite{openrouter2023}: claude-sonnet-4.6, gemini-3.1-flash-lite-preview, gpt-4.1-mini, gemma-4-26b-a4b-it, llama-4-scout, mistral-small-3.2-24b-instruct, qwen3-6-flash, and phi-4, with temperature set to 0.0 for deterministic output. For SFT, 4 encoder-based models reported in recent legal NLP work \cite{singh2025llms} are used: two general-purpose models, BERT and RoBERTa, and two legal-specific models, Legal-BERT and Contracts-BERT. Models are trained with 5 epochs, learning rate 2e-5, batch size 64, and maximum sequence length 256. The SFT results are reported as mean $\pm$ std over 3 seeds. Macro-F1 is used as the evaluation metric, as the dataset is imbalanced and Equivalent is the minority class.
\section{Results and Analysis}
Table \ref{tab_3} reports macro-F1 scores across four settings: AU-UK, UK-IN, IN-AU, and Overall. Country-specific scores represent models trained and evaluated on their respective country pairs, while the overall score represents models trained on all labelled country pairs and evaluated on the whole test set. For AU-UK, the best performance is achieved by CoT using Claude Sonnet 4.6, for UK-IN by Few-Shot using Qwen 3.6 Flash, and for IN-AU by Zero-Shot using Mistral 3.2. Different models and techniques peak on different country pairs, which is expected given the three distinct jurisdictions in LAUKIN. In country-specific settings, prompting-based techniques outperform SFT. However in the overall setting, SFT with BERT achieves the best performance across all techniques and models, suggesting that training on clause pairs from other jurisdictions improves both country-specific and overall performance. Macro-F1 scores vary across country pairs, ranging from 57.75-68.57 for UK-IN, 56.44-66.07 for AU-UK, and 56.21-61.77 for IN-AU. These differences reflect the degree of legal language divergence across jurisdictions. The UK-India pair consistently achieves the highest scores across all models and techniques, as India inherited English legal traditions through British colonial influence, preserving similar clause structures and terminology, making equivalent clauses easier to classify. Although Australia and the UK share a common law heritage, their modern drafting conventions have diverged, resulting in moderate performance on the Australia-UK pair. India-Australia achieves the lowest scores, as both legal systems have evolved independently from their shared colonial origin.
\begin{table}[ht!]
\centering
\caption{LAUKIN test set performance across country pairs and overall. For models trained on the training set, results are reported as mean $\pm$ std. \textbf{Bold} indicates the best-performing model per technique and setting.}
\label{tab_3}
\resizebox{0.49\textwidth}{!}{
\begin{tabular}{ll cccc}
\toprule
\textbf{Technique} & \textbf{Model} 
    & \textbf{AU--UK} & \textbf{UK--IN} & \textbf{IN--AU} & \textbf{Overall} \\
\cmidrule(lr){3-3} \cmidrule(lr){4-4} \cmidrule(lr){5-5} \cmidrule(lr){6-6}
 & & Macro-F1 (500) & Macro-F1 (500) & Macro-F1 (500) & Macro-F1 (1500) \\
\midrule

Zero-Shot & Claude Sonnet 4.6    & 
\textbf{65.41} & 64.61 & 61.28 & 63.9 \\
          & Gemini 3.1 Flash Lite & 62.97 & 62.12 & 60.47 & 62.01 \\
          & GPT-4.1 Mini          & 62.44 & 61.39 & 59.55 & 61.29 \\
          & Gemma 4
       & 63.30 & 65.21 & 59.53 & 62.82 \\
          & Llama 4 Scout         & 63.84 & 60.93 & 60.85 & 62.02 \\
          & Mistral 3.2     & 64.84 & \textbf{65.68} & \textbf{61.77} & \textbf{64.25} \\
          & Qwen 3.6 Flash        & 63.54 & 64.57 & 59.57 & 62.65 \\
          & Phi-4                 & 56.44 & 57.75 & 56.21 & 56.89 \\
\midrule

Few-Shot & Claude Sonnet 4.6    & 64.05 & 65.58 & 60.70 & 63.53 \\
          & Gemini 3.1 Flash Lite & 63.05 & 63.74 & 60.98 & 62.72 \\
          & GPT-4.1 Mini          & \textbf{65.52} & 63.30 & \textbf{61.52} & 63.68 \\
          &Gemma 4
       & 65.07 & 65.00 & 60.46 & 63.75 \\
          & Llama 4 Scout         & 61.97 & 65.52 & 60.44 & 62.75 \\
          & Mistral 3.2     & 61.02 & 64.14 & 58.06 & 61.19\\
          & Qwen 3.6 Flash        & 64.32 & \textbf{68.57} & 60.07 & \textbf{64.35} \\
          & Phi-4                 & 57.11 & 61.74 & 59.05 & 59.39 \\
\midrule

CoT & Claude Sonnet 4.6    & \textbf{66.07} & 64.37 & 60.11 & \textbf{63.66} \\
          & Gemini 3.1 Flash Lite & 62.30 & 62.70 & 59.00 & 61.38 \\
          & GPT-4.1 Mini          & 61.66 & \textbf{65.48} & 58.98 & 62.12 \\
          & Gemma 4 
       &60.04  &64.72  &59.21  &61.29  \\
          & Llama 4 Scout         & 60.81 & 62.26 & 56.51 & 59.95 \\
          & Mistral 3.2     & 63.13 & 61.45 & 59.59 & 61.55 \\
          & Qwen 3.6 Flash        &62.58  &65.26  &\textbf{61.20} &63.04 \\
          & Phi-4                 &60.49  &60.78  &57.45 &59.69  \\
\midrule

SFT       & BERT             & $59.50 \pm_{4.92}$ & $\mathbf{64.43 \pm_{3.37}}$ & $\mathbf{59.00 \pm_{3.87}}$ & $\mathbf{65.11 \pm_{0.20}}$ \\
          & RoBERTa         & $53.74 \pm_{4.86}$ & $56.07 \pm_{6.15}$ & $53.61 \pm_{6.66}$ & $61.31 \pm_{0.24}$ \\
          & Legal BERT & $\mathbf{60.52 \pm_{4.30}}$ & $56.60 \pm_{3.06}$ & $56.18 \pm_{3.80}$ & $62.80 \pm_{0.95}$ \\
          & Contracts BERT   & $59.13 \pm_{5.36}$ & $60.59 \pm_{1.83}$ & $58.61 \pm_{3.62}$ & $64.18 \pm_{0.24}$ \\
\bottomrule
\end{tabular}
}
\end{table}
Table \ref{tab_4} presents full test set performance stratified by word-overlap range, reporting only the best-performing overall model per technique due to space constraints. Two ranges are considered: low (0.0-0.25) and high (0.25-0.63), where 0.63 is the maximum word overlap in the test set (representing 63\% overlap). All best-performing models achieve higher macro-F1 on high word-overlap pairs compared to low word-overlap pairs. SFT with BERT achieves the best overall macro-F1 of 65.11\% (Table \ref{tab_3}), while achieving 71.12\% on high word-overlap pairs and beating all other models and techniques (Table \ref{tab_4}), but scoring 62.07\% on low word-overlap pairs. This indicates that most errors occur on low word-overlap pairs, making equivalence classification difficult. These results demonstrate that LAUKIN is a challenging and competitive benchmark for LLMs.
\begin{table}[ht!]
\centering
\caption{LAUKIN test set performance by word-overlap range for the best model per technique}
\label{tab_4}
\resizebox{0.38\textwidth}{!}{
\begin{tabular}{ll cc}
\toprule
\textbf{Technique} & \textbf{Best Model} 
    & \textbf{0.0-0.25} & \textbf{0.25-0.63} \\
\cmidrule(lr){3-3} \cmidrule(lr){4-4}
 & & Macro-F1 (1391) & Macro-F1 (109)\\
\midrule
Zero-Shot & Mistral 3.2 &           62.60 & 64.33 \\
\midrule
Few-Shot & Qwen 3.6 Flash &        62.53 & 65.64 \\
\midrule
CoT       & Claude Sonnet 4.6 &    61.40 & 67.40 \\
\midrule
SFT       & BERT                   & $62.07 \pm_{0.25}$ & $71.12 \pm_{3.45}$ \\
\bottomrule
\end{tabular}
}
\end{table}
\section{Conclusion}
We introduce LAUKIN, the first multi-jurisdictional common law contract dataset. It comprises 14,727 clause pairs (AU-UK, UK-IN, and IN-AU) from 204 contracts across 8 agreement types. We construct the dataset via a novel multi-stage retrieval and reranking pipeline, with a subset of clause pairs subsequently annotated by legal experts as Equivalent or Not Equivalent. We evaluate 12 models across four techniques. The best macro-F1 of 65.11\% establishes LAUKIN as a challenging benchmark. Prompting-based techniques outperform SFT in country-specific settings; however, SFT with BERT achieves the best overall performance, suggesting that training on clause pairs from other jurisdictions improves performance. Low word-overlap clause pairs remain the primary source of error, posing a significant challenge for legal equivalence classification. LAUKIN also facilitates future semi-supervised learning research. The dataset enables applications including cross-jurisdictional contract review and clause drafting, legal information retrieval, and LLM benchmarking. LAUKIN advances legal NLP beyond single-jurisdiction benchmarks toward real-world cross-jurisdictional legal contract understanding.
\begin{acks}
We thank Praharsh Singh of LexStart Partners for his contributions to the LAUKIN annotation and for his discussions on the legal aspects of this work.
\end{acks}
\section*{GenAI Usage Disclosure}
This paper is written entirely by the authors without any generative AI tools for text generation or writing assistance. All manuscript content, figures, and experimental results are human-produced. AI-based writing assistants are used solely for optional spelling and grammar checks.
\bibliographystyle{ACM-Reference-Format}
\bibliography{sample-base}

\appendix

\end{document}